\documentclass{article}

\usepackage{arxiv}

\usepackage[utf8]{inputenc} 
\usepackage[T1]{fontenc}    
\usepackage{hyperref}       
\usepackage{url}            
\usepackage{booktabs}       
\usepackage{amsfonts}       
\usepackage{nicefrac}       
\usepackage{microtype}      
\usepackage{lipsum}

\usepackage{amsmath,amssymb} 

\usepackage{paralist}
\usepackage[inline,shortlabels]{enumitem}
\usepackage{multirow}
\usepackage{adjustbox}

\title{Weight Pruning via Adaptive Sparsity Loss}

\author{
  George Retsinas\\
  School of ECE, NTUA, Athens, Greece\\
  \texttt{gretsinas@central.ntua.gr} \\
   \And
 Athena Elafrou\\
  School of ECE, NTUA, Athens, Greece\\
  \texttt{athena@cslab.ece.ntua.gr} \\
  \And
 Georgios Goumas\\
  School of ECE, NTUA, Athens, Greece\\
  \texttt{goumas@cslab.ece.ntua.gr} \\
  \And
 Petros Maragos\\
  School of ECE, NTUA, Athens, Greece\\
  \texttt{maragos@cs.ntua.gr} \\
}

\begin{document}
\maketitle

\begin{abstract}
Pruning neural networks has regained interest in recent years as a means to compress state-of-the-art deep neural networks and enable their deployment on resource-constrained devices. 
In this paper, we propose a robust compressive learning framework that efficiently prunes network parameters during training with minimal computational overhead.
We incorporate fast mechanisms to prune individual layers and build upon these to automatically prune the entire network under a user-defined budget constraint.
Key to our end-to-end network pruning approach is the formulation of an intuitive and easy-to-implement adaptive sparsity loss that is used to explicitly control sparsity during training, enabling efficient budget-aware optimization. 
Extensive experiments demonstrate the effectiveness of the proposed framework for image classification on the CIFAR and ImageNet datasets using different architectures, including AlexNet, ResNets and Wide ResNets.
\end{abstract}

\section{Introduction}

Deep neural networks (DNNs) deliver state-of-the-art results in a variety of machine learning tasks, including image classification~\cite{NIPS2012_4824}, visual object recognition~\cite{Ren:2017:FRT:3101720.3101780}, speech recognition~\cite{NIPS2015_5847} and natural language processing~\cite{NIPS2014_5346}.
This has lead to an increasing adoption of deep learning as the central machine learning technology behind a wide variety of applications in modern society. 
Many of these applications are deployed on mobile, resource-constrained devices and cannot afford to offload computations to cloud computing infrastructure due to latency or privacy restrictions. 
Such devices, however, do not typically meet the high computational and memory demands of state-of-the-art DNNs and, thus, the quality of these applications heavily depends on building compact, yet accurate models.
This has triggered an extensive line of research on efficient network design \cite{Chollet_2017_CVPR,He_2016_CVPR,Howard2017MobileNetsEC,Xie_2017_CVPR,BMVC2016_87,Zhang_2018_CVPR} and model compression via pruning~\cite{8237417,8237803,NIPS2016_6504}, quantization~\cite{NIPS2016_6573,Jacob_2018_CVPR}, tensor decomposition~\cite{NIPS2013_5025,NIPS2014_5544,Liu_2015_CVPR} and knowledge distillation~\cite{44873}.

Network pruning has shown promising results in achieving high compression rates with minimal accuracy loss~\cite{NIPS2016_6165,han2015deep,NIPS2015_5784}.
The main assumption behind this method is that DNNs are often over-parameterized and, thus, one can obtain comparable accuracy using a subset of the trained parameters. 
Depending on the granularity of parameters that are removed, pruning methods can be generally categorized as \emph{unstructured} or \emph{structured}. Unstructured pruning---also referred to as \emph{weight pruning}---removes individual parameters (connections), generating a sparse model that preserves the high-dimensional features of the original network.
On the other hand, structured pruning removes individual channels/filters (neurons) producing a thinner, denser model.
From a theoretical standpoint, unstructured pruning can preserve more accuracy under coarser compression rates. It does, however, require optimized implementations of sparse tensor operations~\cite{Li:2018:HHS:3291656.3291682} to translate the memory and computational savings to practical performance gains on commodity hardware.
Nonetheless, many hardware accelerators for DNNs have been proposed with support for unstructured sparsity as their key design goal~\cite{DelmasLascorz:2019:BSA:3297858.3304041,Gondimalla:2019:SST:3352460.3358291,Hegde:2019:EAS:3352460.3358275,Lu:2018:SES:3195970.3196120,Parashar:2017:SAC:3079856.3080254,You:2019:RAS:3289602.3293945,Zeng:2019:FSA:3289602.3293964,Zhou:2018:CAI:3343260.3343263,Zhu:2019:STC:3352460.3358269}.

Previous work on weight pruning attempts to sparsify a network either \emph{indirectly} by adding a sparsity inducing penalty term in the learning process~\cite{Carreira-Perpinan_2018_CVPR,DBLP:journals/corr/abs-1905-13678,Molchanov:2017:VDS:3305890.3305939,10.5555/3042817.3043055} or \emph{directly} by removing weights from a fully-trained reference network based on a ``saliency'' ranking criterion~\cite{NIPS2016_6165,han2015deep,NIPS1992_647,NIPS1989_250}.
In the first case, heavily-tuned hyper-parameter settings are often required to obtain high sparsity levels.
In the latter case, it is necessary to perform multiple prune-retrain cycles---commonly referred to as fine-tuning---to minimize accuracy loss, which is typically very time-consuming and can, thus, hinder their application on large models.

In this work, we build low-overhead, efficient weight pruning mechanisms directly into the learning process.
First, we propose a methodology to remove unimportant weights in every training iteration under user-defined sparsity constraints using fast and effective magnitude-based pruning functions on individual layers (Section~\ref{sec:fixed}).
Then, we extend this methodology so that the per-layer sparsities can also be discovered automatically by introducing a novel adaptive sparsity loss (Section~\ref{sec:adaptive}).

For pruning individual layers, we explore two cost-effective magnitude-based threshold functions: the first sets the pruning threshold using a binary search algorithm, while the second uses confidence interval analysis under a Gaussian assumption for the distribution of weights.
Both variations aim to retain a certain predefined sparsity level.
Motivated by previous work~\cite{Carreira-Perpinan_2018_CVPR,NIPS2016_6165}, we also allow pruned connections to be recovered in subsequent iterations.
The key enabling factor to fast and efficient recovery of previously pruned connections is a novel application of the Straight Through Estimator (STE)~\cite{DBLP:journals/corr/BengioLC13}, which is used to update weights during the back-propagation step.
STE enables weight re-use while preserving a bell-shaped weight distribution, which is crucial for effectively removing connections using the proposed pruning functions.

Unlike methods that rely on complicated sparsity inducing regularisers~\cite{Carreira-Perpinan_2018_CVPR,DBLP:journals/corr/abs-1905-13678,Molchanov:2017:VDS:3305890.3305939,10.5555/3042817.3043055}, we introduce an intuitive sparsity controlling loss, which allows one to tune the overall size-accuracy trade-off of the model depending on the target platform and application requirements.
Specifically, under a Gaussian assumption for the weight distributions, we formulate the sparsity level of each layer as the error function (\emph{erf}) depending on the threshold value.
This function is differentiable and, thus, back-propagatable. 
In this way, we define the overall sparsity of the network by combining the per-layer sparsities into an extra loss function to be optimized along with the task-related loss.
Eventually, each layer's threshold is a trainable parameter controlled by the sparsity function and we can combine sparsities into a budget-aware fashion, enforcing an overall budget on the number of parameters or operations and granting the method freedom to prune layers that are more redundant more aggressively.

Overall, the proposed weight pruning framework has several compelling properties:
\begin{itemize}
    \item \emph{Trainable sparsity}: optimal per-layer sparsities are automatically determined under a user-defined budget constraint.
    \item \emph{Versatility}: our methods are robust to different layer types (convolutional, fully connected etc.) and architectures. Moreover, the proposed adaptive sparsity loss can be easily modified to match the user's needs.
    \item \emph{Minimal computational overhead:} efficient pruning is achieved with negligible convergence overhead.
\end{itemize}

Experimental results on the CIFAR-100~\cite{krizhevsky2009learning} and ImageNet~\cite{ILSVRC15} datasets with a variety of network architectures indicate the effectiveness of the proposed methods on image classification
(Section~\ref{sec:experimental}). 
Driven by the lack of a consistent methodology for evaluating pruning methods, we also advocate \begin{enumerate*}[(a)]
    \item using a dense equivalent network with the same budget as a comparison point and
    \item creating trade-off curves between the compression rate and classification error
\end{enumerate*}
(Section~\ref{sec:fair_comparison}). 


\section{Related Work}

Since the advent of AlexNet~\cite{NIPS2012_4824}, the first DNN to achieve state-of-the-art accuracy in the highly challenging ImageNet classification task~\cite{ILSVRC15}, building deeper and wider neural networks has been the main trend towards improving performance in a variety of machine learning tasks.
While the requirements of modern DNNs can easily be met by server-class computing systems, this is not the case for less powerful mobile or embedded devices, thus limiting the deployment of state-of-the-art models in many real-life applications.
This has motivated researches to follow two main directions: 
\begin{enumerate*}[(a)] 
    \item design compact, yet well-performing architectures and 
    \item develop methods to compress pre-trained models with minimal loss in accuracy.
\end{enumerate*}

\textbf{Compact Architectures.} 
Many architectural improvements have been devised to improve the cost efficiency of CNNs either by replacing a costly convolutional layer with a set of cheaper convolutions (e.g. pointwise or grouped convolutions)~\cite{Chollet_2017_CVPR,Han_2017_CVPR,He_2016_CVPR,Howard2017MobileNetsEC,Xie_2017_CVPR,BMVC2016_87,Zhang_2018_CVPR} or by creating complex convolution flows (e.g. multi-branched convolutional blocks) instead of sequentially convolving into a single flow~\cite{He_2016_CVPR,huang2017densely,larsson2016fractalnet,retsinas2019recnets,Szegedy_2015_CVPR}.

\textbf{Model Compression.} Driven by the observation that deep learning models tend to be hugely over-parameterized~\cite{NIPS2013_5025}, researchers have also explored different ways of compressing pre-trained models while preserving accuracy.
\emph{Network pruning}, pioneered in the early development of neural networks~\cite{NIPS1989_250}, seeks to induce sparsity in a neural network by removing unimportant connections. Different approaches vary in terms of the granularity at which they prune the network and the criteria used to determine connection importance. Researchers have proposed pruning individual weights~\cite{NIPS2017_6910,Carreira-Perpinan_2018_CVPR,DBLP:journals/corr/abs-1905-13678,NIPS2016_6165,Molchanov:2017:VDS:3305890.3305939} or channels/filters~\cite{Ding_2019_CVPR,Gordon_2018_CVPR,8237417,He_2019_CVPR,8237803,Molchanov_2019_CVPR,NIPS2016_6504,Yang_2018_ECCV,Zhao_2019_CVPR,NIPS2018_7367}.
\emph{Quantization}, which dates back to the 1990s~\cite{fiesler1990weight}, uses low-precision fixed-point instead of high-precision floating-point formats to represent the weights and/or activations of a model in order to reduce its memory footprint both in terms of storage and working memory during inference~\cite{Cai_2017_CVPR,NIPS2015_5647,Hubara:2017:QNN:3122009.3242044,10.1007/978-3-319-46493-0_32}.
Other approaches include \emph{tensor decomposition} methods~\cite{NIPS2013_5025,NIPS2014_5544,Liu_2015_CVPR} which derive low-rank approximations of the weight tensors and \emph{knowledge distillation} which transfers knowledge from a over-parameterized teacher network to a smaller student network~\cite{Bucilua:2006:MC:1150402.1150464,NIPS2018_7553,44873}. 

\section{Online Weight Pruning}
\label{sec:proposed}

\subsection{Preliminaries}

Given a layer comprising a weight tensor $\mathbf{W}$ and a bias tensor $\mathbf{b}$, we focus on sparsifying $\mathbf{W}$, since the bias term has a minor contribution to its budget.
Pruning individual weights directly affects the size of the layer (and consequently the model) and the number of floating-point operations required to infer. Specifically, both layer characteristics (size and required operations) are scaled linearly by the density percentage, i.e. $1-s$ if $s$ is the layer's sparsity.
In the following, we assume a neural network consisting of $N$ such layers and even though the notation describes convolutional layers, it can be straightforwardly applied to any type of trainable layer.

\subsection{Magnitude-based Pruning}

Using magnitudes to determine the saliency of connections in neural networks is prominent in both structured and unstructured pruning techniques. 
The reasoning is simple and intuitive: small magnitude indicates small contribution to the output.
For the case of weight (unstructured) pruning, the simplest approach is to remove individual weights $w$ according to their absolute value given a bound $b$ as follows:
\begin{equation}
f_{prune}(w; b) =
    \begin{cases}
      0, & \text{if}\ |w| < b \\
      w, & \text{otherwise}
    \end{cases}
\label{eq:condition}
\end{equation}

\begin{figure}[t]
\centering
\begin{tabular}[t]{cc}
\begin{adjustbox}{valign=c}
\begin{tabular}{cc}
\includegraphics[scale=0.2]{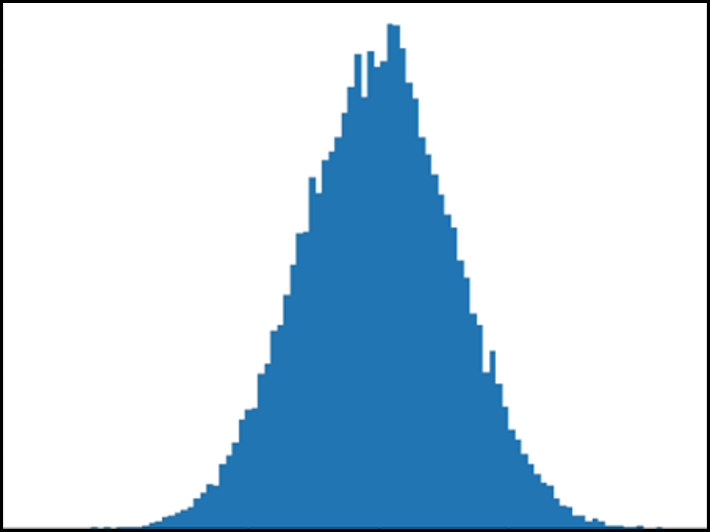} & \quad
\includegraphics[scale=0.2]{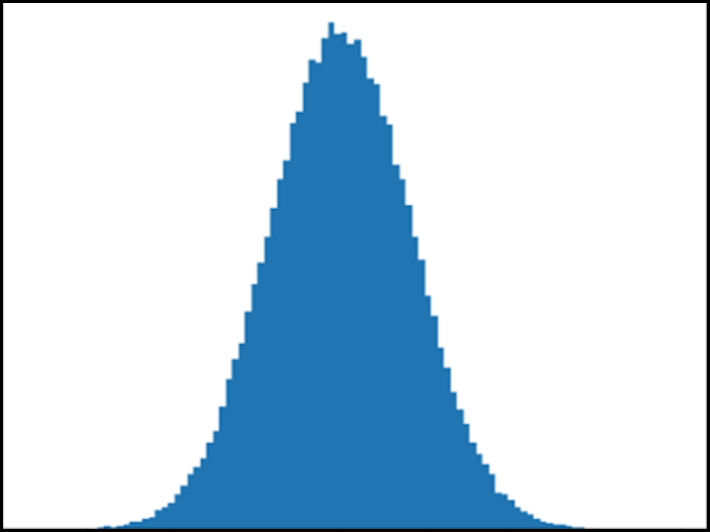}
\\
\includegraphics[scale=0.2]{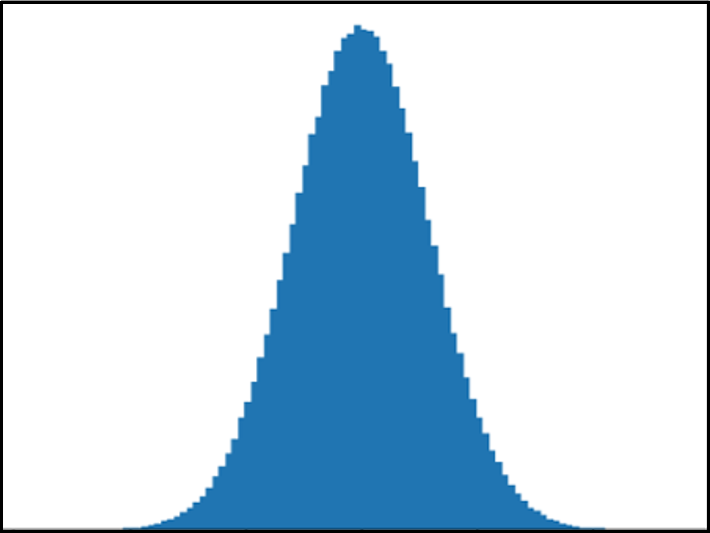} & \quad
\includegraphics[scale=0.2]{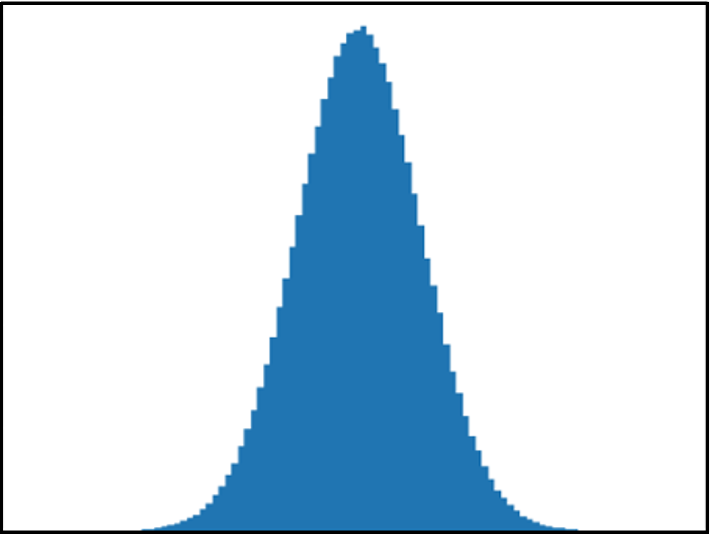}
\end{tabular}
\end{adjustbox}
& \quad\quad
\begin{adjustbox}{valign=c}
\includegraphics[scale=0.41]{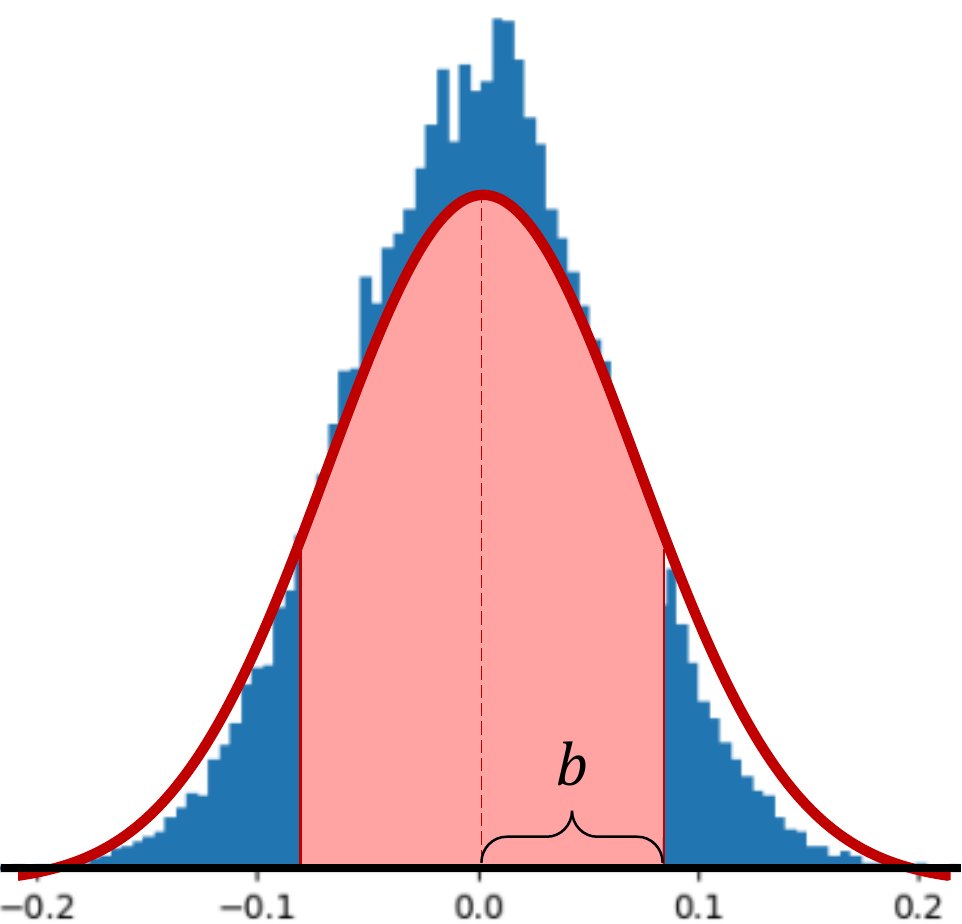}
\end{adjustbox}\\
(a) & \quad\quad (b)
\end{tabular}
\caption{(a) Distribution of weights of the last convolutional layer in each block of ResNet50 trained on the ImageNet dataset. (b) Example of a Gaussian bell over the real distribution of weights (layer2\_1 of ResNet50).}
\label{fig:distr}
\end{figure}

Despite the simplicity of this approach, it has been shown to be effective, leading to significant compression rates~\cite{NIPS2016_6165}.
Key to its effectiveness is the assumption of a unimodal unknown weight distribution, centered at zero.
In practice, weight decay~\cite{NIPS1991_563}, which is typically used in training, imposes the minimization of the parameters' values or, in other words, the concentration of the parameters' values around zero, and, thus, favors this assumption.    
Indeed, Figure~\ref{fig:distr}a verifies this assumption for layers of a fully-trained ResNet50 model. 
For the sake of brevity, we henceforth assume a zero-mean weight distribution per layer. Nonetheless, the following analysis can be easily adapted for a non-zero mean without loss of generality.

Following the pruning condition of Equation~\ref{eq:condition}, we formulate the problem of weight pruning as selecting the appropriate bound $b$ per layer to achieve a target sparsity level with minimal accuracy loss. 
In online pruning, imposing a fixed value for the bound is destined to fail since the weights are continuously updated and, eventually, the vast majority will end up outside the bounded interval, resulting in an almost entirely dense layer. 
Therefore, dynamic bounds need to be considered.
Specifically, the bound $b$ should depend on the weights $\mathbf{W}$, i.e. $b = f_b(\mathbf{W})$.
In the following sections, we use such dynamic bounds in order to develop efficient weight pruning approaches under different constraints.

\subsection{Fixed Sparsity}
\label{sec:fixed}

First, we aim to achieve a predefined level of sparsity per layer. 
Nonetheless, it is straightforward to define an overall sparsity for the neural network by requesting the same sparsity level for every layer.
Following the discussed hypothesis of a unimodal weight distribution, given a predefined sparsity level $s$, we seek to find an appropriate bound function $f_b(\mathbf{W};s)$.
We follow two approaches for this task:
\begin{enumerate*}[(a)]
\item find $b$ using a binary search algorithm and
\item assume a normal distribution over the weights and find $b$ using a confidence interval analysis.
\end{enumerate*}
The trade-off between the two approaches lies in the sparsity precision versus the computational overhead.
Specifically, the first approach is more precise but more expensive, while the second is faster to compute but is based on a rather weak assumption (normal distribution) that may lead to deviations from the requested sparsity. 

\subsubsection{Binary Search}
\label{sec:fixed:binary}

Given an approximation margin $\epsilon$ and a sparsity level $s$, the problem of finding an appropriate bound $b$ can be formulated as follows:
\begin{align}
\left| \frac{\#\{|w| < b \, | \, w \in \mathbf{W}\}}{\#\mathbf{W}} - s \right| < \epsilon
\label{eq:bs}
\end{align}
The above problem formulation can be efficiently solved w.r.t. $b$ via binary search.
At each step of the binary search algorithm we need to re-estimate the bound $b$ and, thus, we have to count the newly pruned weights. 
The total number of steps depends on the approximation parameter $e$.
Contrary to using a sorting algorithm for this task, the binary search approach provides a faster convergence to bound $b$ under the approximation error $\epsilon$, which is crucial when considering layers comprising millions of parameters. 
As we will discuss further on, this approach serves as a \emph{baseline}.

\subsubsection{Gaussian Assumption}
\label{sec:fixed:gauss}

The per-layer weight distribution often resembles a Gaussian bell curve (e.g., see Figure~\ref{fig:distr}a). 
This observation leads to the idea of selecting a statistical-related pruning bound according to a confidence interval analysis, by requesting a confidence level equal to the target sparsity level, as shown in Figure~\ref{fig:distr}b. 
Similar statistical approximation of $b$ has been previously examined at~\cite{xu2018hybrid} by introducing a sparsity level controlling factor $a$ and computing the bound according to: $b = a \times \text{std}(\text{vec}(\mathbf{W}))$. 
Nonetheless, the authors of~\cite{xu2018hybrid} do not explore the ability to precisely define sparsity via an appropriate $a$ based on confidence intervals.

More formally, assuming a layer's weights follow a zero-centered normal distribution, i.e. $w \sim N(0, \sigma)$, and given a desired level of sparsity $s\in(0,1)$, the pruning bound $b$ can be computed as:
\begin{align}
b = f_b(\sigma; s) = \sigma \sqrt{2} \, \text{erf}^{-1}(s) \label{eq:ga}\\
\text{where} \quad \text{erf}(x) = \frac{1}{\sqrt{\pi}}\int_{-x}^{x}e^{-t^2}dt
\end{align}
In practice, we only need to compute the standard deviation $\sigma$ of the weights in order to find an appropriate $b$.
It is important to note that the Gaussian assumption is fairly weak (see Figure~\ref{fig:distr}b).
Therefore, it is possible to encounter deviations between the requested and attained sparsity levels. 
If the sparsity level exceeds the target, there may be a negative impact on the attainable accuracy. If it is lower, it may violate the user's requirements.

\subsubsection{Training using a Straight-Through Estimator}

The efficiency of the proposed methodology relies on a simple, yet powerful concept, referred to as Straight-Through Estimator (STE)~\cite{DBLP:journals/corr/BengioLC13}.
STE was initially motivated by the need for a differentiable function for threshold operations in order to properly train a network.
To overcome this problem, Hinton et al. \cite{DBLP:journals/corr/BengioLC13} proposed to back-propagate through such hard threshold functions as if it had been the identity function (hence the straight-through term).
Considering the problem at hand, the main idea is to update all weights, as if they actively contribute to the layer output, via the back-propagation scheme of STE.
However, during the forward propagation, we only use the sparse subset of weights in order to generate the layer output.
In other words, forward and backward propagation steps (with respect to the weight tensor) are mutually inconsistent.
This inconsistency is very helpful; the network is trained according to the sparsity constraints, while simultaneously all weights, even if they are pruned, are updated in order to optimize the layer output. 
In this way, an already pruned weight may return to the used weight set without any complications since it has already been trained all along.

Let $\mathbf{X}$ and $\mathbf{Y}$ be the input and output of a layer, respectively, $\mathbf{W}$ be the weight tensor of the layer and 
$\widetilde{\mathbf{W}} = f_{prune}(\mathbf{W})$ be the pruned weight tensor. 
Given a loss function $L$ at the output of the network, the aforementioned approach for a given convolutional layer is formulated as follows ($*$ and $*^\intercal$ are the convolution and the transposed convolution operations, respectively):

\begin{itemize}
    \item \emph{Forward:} $\mathbf{Y} = \mathbf{X} * \widetilde{\mathbf{W}}$
    \item \emph{Backward:}
    $\dfrac{\vartheta L}{\vartheta \mathbf{X}}  = \dfrac{\vartheta L}{\vartheta \mathbf{Y}} *^\intercal \widetilde{\mathbf{W}} \, , \quad
    \dfrac{\vartheta L}{\vartheta \mathbf{W}}  = \mathbf{X} * \dfrac{\vartheta L}{\vartheta \mathbf{Y}}$
    \item \emph{Update:} $\mathbf{W} \gets \mathbf{W} - \eta \dfrac{\vartheta L}{\vartheta \mathbf{W}} $
\end{itemize}

\begin{figure}[t]
\centering
\includegraphics[width=0.42\linewidth]{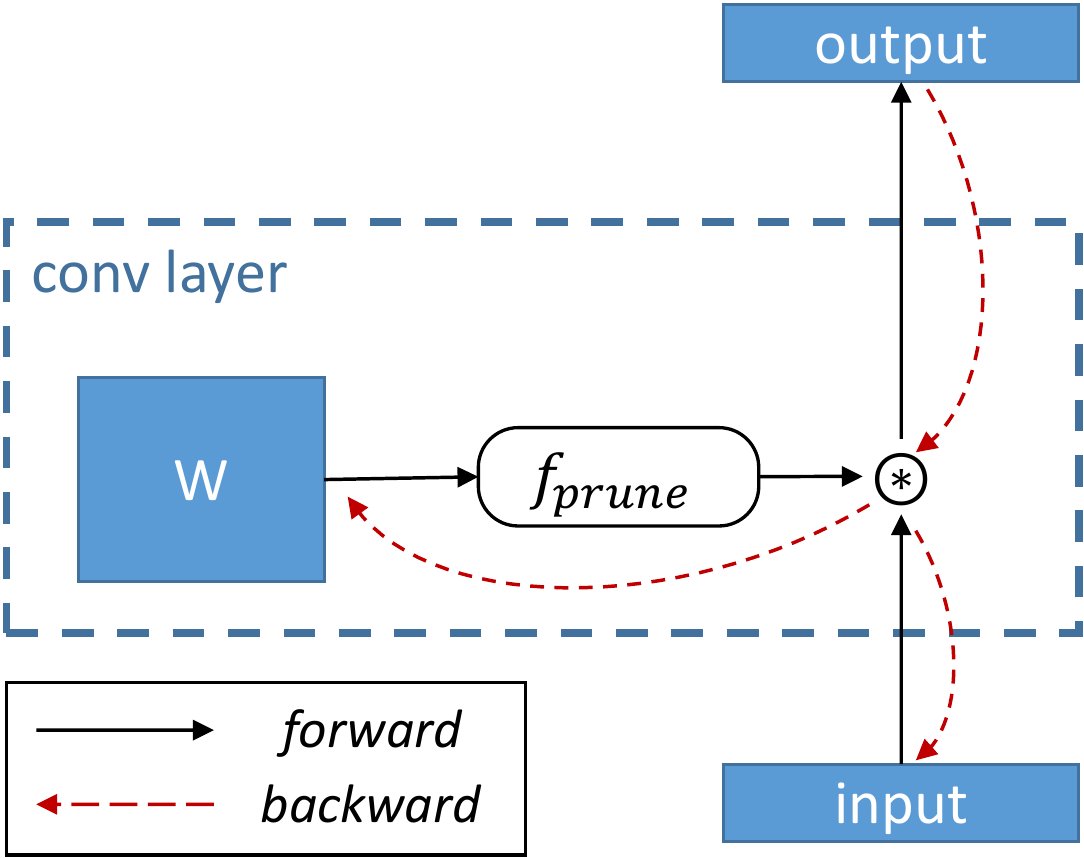}
\caption{Online pruning using Straight-Through Estimator for updating weights. The pruning operation is bypassed during the back-propagation pass (see red dotted arrows) and thus every weight is updated.}
\label{fig:ste}
\end{figure}

The functionality of STE in the context of our pruning framework is depicted in Figure~\ref{fig:ste}. 

\begin{figure}[t]
\centering
\begin{tabular}[t]{cc}
\includegraphics[width=0.26\linewidth]{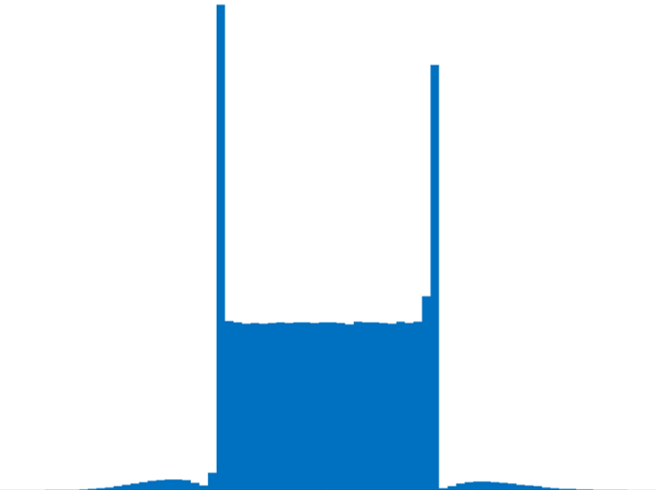} & 
\includegraphics[width=0.26\linewidth]{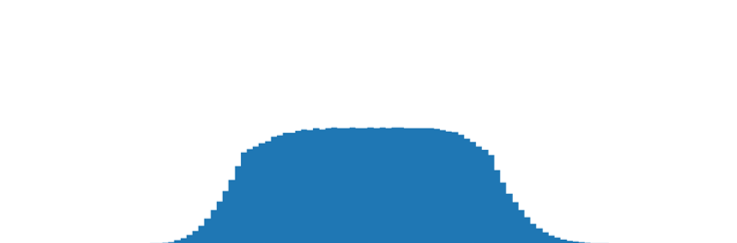} \\
(a) w/o STE & (b) w/ STE
\end{tabular}
\caption{Impact of STE on weight distribution; STE preserves a bell-shaped weight distribution.}
\label{fig:ste-distr}
\end{figure}

The main motivation behind using STE in our work is to enable weight re-use and preserve a bell-shaped distribution, as opposed to the initial motivation of meaningful derivatives. 
Indeed, the hard-shrink function of Eq.~\ref{eq:condition} can be back-propagated without complications.
Nevertheless, back-propagating through this function would exclude the pruned weights from the update step, resulting in non-unimodal distributions. 
An example of this phenomenon is depicted in Figure~\ref{fig:ste-distr}a, where the non-pruned weights form two small areas around the main pruned area.
On the contrary, using STE leads to a well-formed bell-shaped weight distribution as Figure~\ref{fig:ste-distr}b suggests.
Therefore, STE is crucial to the efficiency of the second bound selection approach (see Section~\ref{sec:fixed:gauss}) which makes a Gaussian assumption of the distribution of weights.
Indeed, using this bound selection approach without STE under a sparsity target of $85\%$ results in $93.2\%$ sparsity---a significant deviation---while enabling STE leads to a sparsity of $85.6\%$, proving the necessity of STE when considering the Gaussian assumption approach. Note that our goal is to obtain sparse models according to a sparsity target and thus such deviations are not desirable, while higher sparsity (above a specific threshold -- see Section~\ref{sec:fair_comparison}) may lead to a significant drop in performance.  

Updating the entire weight tensor for dynamic weight re-usage is not a new concept (e.g. \cite{NIPS2016_6165}), even though it has not been formulated as a use case of STE before and, more importantly, it has not been connected to the distribution preservation key property.  



Apart from STE, we also adopt the $L^2$ regularization term, also referred to as weight decay~\cite{NIPS1991_563}, which has been shown to provide solutions that generalize well. Overall, the use of STE along with weight decay aims to assist the convergence on a subset of weights with good generalization properties and a bell-shaped distribution, while the pruning function is responsible for sparsifying the network.
A key advantage of this approach is that it enables low-overhead online pruning, as opposed to the typical three-stage pipeline consisting of first fully-training the reference network, then pruning and fine-tuning to recover accuracy\footnote{The last two steps may be performed once, often referred to as \emph{one-shot} pruning, or repeated, often referred to as \emph{iterative} pruning.}~\cite{NIPS2017_6910,8237417,He_2019_CVPR,Molchanov_2019_CVPR}.

\subsection{Adaptive Sparsity}
\label{sec:adaptive}

The pruning methods presented in the previous section require the user to define the level of sparsity per layer. 
While a straightforward solution would be to request the same sparsity per layer, it may be difficult to achieve the desired compression/accuracy trade-off.
To overcome this issue, in this section we propose a novel adaptive sparsity loss, capable of automatically pruning a network with variable sparsity per layer given an overall sparsity constraint. 
In order to define such a loss, a differentiable loss function is required.
Whilst the binary search variation is more precise, only the Gaussian variation can be described by an analytical differentiable function, ideal for  back-propagating a sparsity loss.
Similarly to fixed sparsity pruning, training with a sparsity loss heavily relies on STE for preserving the desired properties of the weights' distribution and retaining a competitive convergence rate.


\subsubsection{Sparsity Percentage as a Loss Function}

As the analysis of \ref{sec:fixed:gauss} hints, the sparsity percentage $s$ of a layer can be defined as a function of the adaptive threshold $b$ and the standard deviation of weights $\sigma$ as follows: 
\begin{equation}
s = \text{erf}(\frac{b}{\sigma \sqrt{2}})
\end{equation}
The key observation here is that the $\text{erf}()$ function is differentiable and can be used to define a simple yet intuitive extra loss term with the aim to find the best possible threshold $b$ that maximizes the sparsity (or minimizes the density) of a particular layer.
To this end, a straightforward solution is to define the average network density as the following loss:
\begin{align}
    L_s(\{b_i\}) & = \frac{1}{N} \sum_i^N (1-s_i) = 1 - \frac{1}{N} \sum_i^N \text{erf}\Big(\frac{b_i}{\sigma_i \sqrt{2}}\Big)
\label{eq:avg_sparsity}
\end{align}

At the same time, we aim to preserve the network accuracy, thus the overall loss is defined as a multitask problem:
\begin{equation}
    L(\{\textbf{W}_{i}\}, \{b_i\}) + \lambda L_s(\{b_i\})
\label{eq:mloss}
\end{equation}
where $\textbf{W}_i$ is the weight tensor and $b_i$ the boundary of the $i^{th}$ layer.
The term $L$ corresponds to the task-related loss (e.g. classification) and, as the above formulation suggests, our approach is independent of the main loss and can be used with any architecture and loss function.

As Eq.~\ref{eq:mloss} suggests, we minimize the loss with respect to both the set of weight tensors $\{\textbf{W}_{i}\}$ and the set of boundaries $\{b_i\}$. 
This means that for every layer we add a single \emph{trainable} parameter $b_i$, resulting in a trivial parameter overhead ($N$ in total).  
The parameter $b_i$, apart from contributing to the sparsity loss, affects the hard-shrink operation that is responsible for weight pruning\footnote{In practice, at each layer, we formulate the pruning operation as: $\widetilde{\mathbf{W_i}} = b_i \cdot \sigma_i \cdot f_{prune}(\mathbf{W_i}/(b_i \cdot \sigma_i); 1)$.}. 
Therefore, if we optimize freely over the $L(\{\textbf{W}_{i}\}, \{b_i\})$ term, i.e. without the sparsity constraint, the boundaries $\{b_i\}$ would be minimized in order to leave the majority of weights un-pruned and subsequently achieve high accuracy.
In other words, the $L$ and $L_{s}$ terms act in a competitive manner.
The back-propagation step and the derivatives according to the STE formulation are provided in detail in the appendix.
The user defined hyper-parameter $\lambda$ controls the contribution of the sparsity loss term.
Note that for a set of tasks (e.g. find optimum sparsity with minimal impact on accuracy) it is easier to set the sparsity loss contribution than the level of sparsity.

\subsubsection{Size-aware Adaptive Sparsity}

Eq. 6 makes no assumptions on the contribution of each layer to the overall model accuracy and, thus, attempts to equally sparsify (percentage-wise) every layer in the network.
Depending on the network architecture and target dataset, however, this may not be ideal. It may be the case, for instance, that layers with more parameters should be pruned more aggressively.
This concept can be easily translated into a loss of weighted summation of the $\text{erf}()$ functions, i.e.:
\begin{equation}
    L_s(\{b_i\};\{c_i\}) = 1 - \frac{1}{N}\sum_i^N c_i \text{erf}\Big(\frac{b_i}{\sigma_i \sqrt{2}}\Big)
\label{eq:wavg}
\end{equation}
where $c_i$ is the per-layer contribution weight. 
For example, in order to minimize the number of parameters in the network, we could define the contribution weights as: 
\begin{equation}
   c_i = \frac{\#\textbf{W}_i}{\sum_j \#\textbf{W}_j}
\label{eq:cweights}
\end{equation}
Similarly, we can define contribution weights with respect to flops per layer in order to construct a flops-centered minimization loss.

\subsubsection{Budget-constrained Adaptive Sparsity}

Finally, we also examine the case of budget-constrained optimization by transforming the overall loss as follows:
\begin{equation}
    L_{CE}(\{\textbf{W}_i\}, \{b_i\}) \, + \,  \lambda_p \Vert L_s(\{b_i\};\{c_{p,i}\}) - B_p \Vert^2\\ \, + \, \lambda_f \Vert L_s(\{b_i\};\{c_{f,i}\}) - B_f \Vert^2
\label{eq:budget}
\end{equation}
where $B_p$ and $B_f$ are the budget percentage for the number of parameters and flops respectively, $c_{p,i}$ and $c_{f,i}$ are the contribution weights per layer (defined as in Eq.~\ref{eq:cweights}), while $\lambda_p$ and $\lambda_f$ are user-defined hyper-parameters.

Alternatively, budget-constrained optimization can be modeled to prohibit the acquired budget, after the sparsification, to exceed the requested target budget. This approach favors sparser models, if possible, and is formulated as: 
\begin{equation}
    L_{CE}(\{\textbf{W}_i\}, \{b_i\}) \, + \,  \lambda_p \max(L_s(\{b_i\};\{c_{p,i}\}) - B_p, 0) \, + \, \lambda_f \max(L_s(\{b_i\};\{c_{f,i}\}) - B_f, 0)
\label{eq:budget2}
\end{equation}

\subsubsection{Possible Extensions}

The proposed adaptive sparsity loss is modular and can, in fact, be combined with any task-specific loss and neural network type, e.g.\ convolutional, recurrent, etc. 
Apart from using any type of network and task, the user can also formulate the sparsity constraints as an extra loss according to the problem at hand.
For example, one may require to either have very sparse or very dense layers, while  approximating an overall target sparsity $s_t$, i.e.\ $\sum_i^N s_i \simeq s_t$, while $s_i < s_l$ or $s_i > s_u$, where $s_l$ and $s_u$ are the lower and upper thresholds respectively.


\section{Evaluation of Pruning: Trade-off Curves}
\label{sec:fair_comparison}

\begin{figure}[t]
\begin{center}
\includegraphics[width=0.55\linewidth]{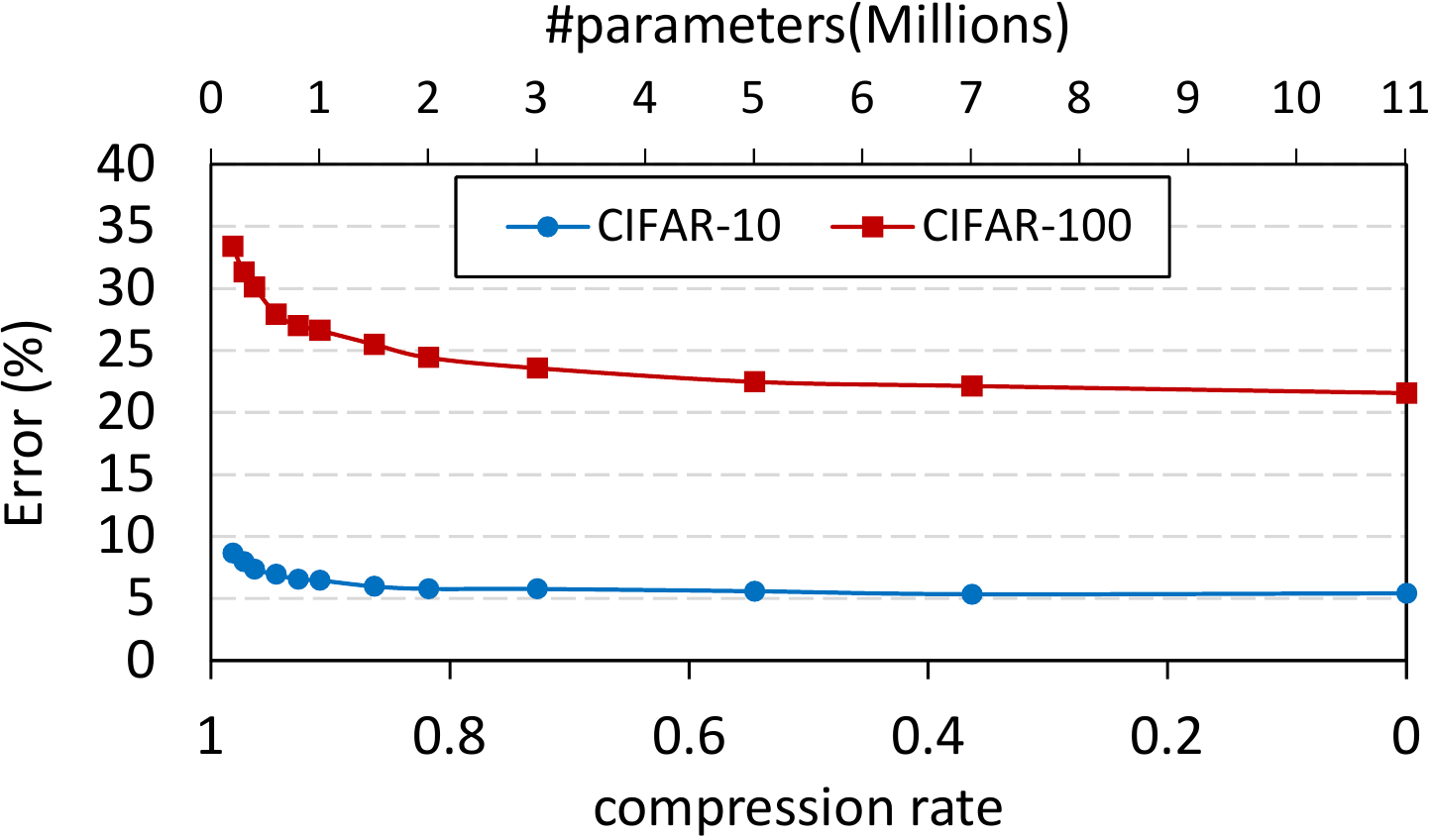}
\end{center}
\caption{Error vs parameter budget trade-off curve for the CIFAR datasets. The initial network was WRN-16-8~\cite{BMVC2016_87}.}
\label{fig:curve}
\end{figure}

Simply reporting the compression ratio vs accuracy loss trade-off is insufficient to validate the effectiveness of pruning or, in general, any compression technique, as the question that often arises is whether the same level of accuracy can be actually attained by training a thinner network with the same resource budget. 
This can be especially true when pruning highly over-parameterized models.
In order to answer this question, we need to compare the accuracy of the compressed network to that of a thinner version of the original network with the same resource budget.
Here, we consider the number of parameters as our budget (analysis for flops is very similar).
Specifically, we use the same architecture and layer depth and reduce the number of channels by the same percentage at each layer so that the size of the thinner dense model matches the size of the compressed model.
If $C'_{in}$/$C'_{out}$ are the number of input/output channels of a layer in the thin model, $C_{in}$/$C_{out}$ are the initial input/output channels of the dense model and $r_c$ is the overall compression ratio, then:
\begin{align}
C'_{in} = \lfloor \sqrt{r_c}\, C_{in} \rfloor, \quad C'_{out} = \lfloor \sqrt{r_c}\, C_{out} \rfloor
\end{align}
Based on this analysis, we can build a trade-off curve reporting the error versus the parameter budget for a specific architecture and dataset, as shown in Figure~\ref{fig:curve}. 
The plotted curve indicates that the WRN-16-8 network is over-parameterized for the CIFAR datasets and, thus, we can successfully train much thinner versions with considerably fewer parameters without sacrificing accuracy.
For larger datasets and architectures, plotting such a curve can be rather impractical despite being insightful.
Nevertheless, in order to objectively evaluate a compressed network, one should report the accuracy of an equivalent dense model with the same budget.

\section{Experimental Results}
\label{sec:experimental}


In this section, we evaluate the proposed pruning methods\footnote{The code is publicly available at \url{https://github.com/georgeretsi/SparsityLoss}.} for the task of image classification on two popular datasets, namely CIFAR-100~\cite{krizhevsky2009learning} and ImageNet~\cite{ILSVRC15}.
Although CIFAR-10 is also a popular choice, we see no merit in exploring such an extreme case of over-parameterization, as demonstrated in Figure~\ref{fig:curve} where re-training a thinner network from scratch has a minor effect to the overall performance.
The considered architectures are the widely-used ResNets~\cite{He_2016_CVPR} and Wide ResNets~\cite{DBLP:journals/corr/ZagoruykoK16}, as well as AlexNet, which is commonly used in the literature to evaluate weight pruning methods.   

\subsection{CIFAR-100 Exploration}


We evaluate the proposed methods on CIFAR-100 using the Wide ResNet architecture and specifically the WRN-16-8 model ($\sim$11 million parameters).
We employ a SGD optimizer for 120 epochs and a cosine annealing scheduling with restarts every 40 epochs~\cite{loshchilov2016sgdr}.


The proposed pruning variations are summarized in Table~\ref{table:proposed}, where we distinguish two main categories: 1) fixed target sparsity per layer and 2) adaptive sparsity per layer. The latter relies on the proposed sparsity loss as described in Section~\ref{sec:adaptive} and can be categorized into unconstrained minimization of the adaptive sparsity loss, i.e. seeking the optimal sparsity without sacrificing accuracy, and budget-constrained optimization with respect to a predefined level of overall sparsity. 
Considering each layer's contribution to the overall sparsity loss, both adaptive techniques can be used with (\emph{wavg}) or without (\emph{avg}) the layers' contribution weights $c_i$ of Eq.~\ref{eq:cweights}.

We evaluate the case of budget-constrained optimization (Eq.~\ref{eq:budget} with $\lambda_f=0$) with an overall requested sparsity set to 85\% at Table~\ref{table:cifar100}, reporting performance of both the initial WRN-16-8 and the compressed dense equivalent WRN-16-3. Since a budget is set, the \emph{unconstr} approach is not applicable.

\setlength{\tabcolsep}{4pt}
\begin{table}[h]
\centering
\caption{Proposed sparsity-controlling variations}
\label{table:proposed}
\begin{tabular}{p{.28\linewidth}p{.15\linewidth}p{.5\linewidth}}
\hline\noalign{\smallskip}
Sparsity & Approach & Description \\
\noalign{\smallskip}
\hline
\noalign{\smallskip}
\multirow[t]{2}{*}{fixed (layer-wise)} & BS & Binary Search (Eq.~\ref{eq:bs}) -- \emph{baseline}.\\
 & GA & Gaussian Assumption (Eq.~\ref{eq:ga}).\\
\hline
\multirow[t]{4}{*}{adaptive (network-wise)} & \multirow[t]{2}{*}{unconstr} & Unconstrained minimization of the overall sparsity loss (Eq.~\ref{eq:wavg}).\\
& \multirow[t]{2}{*}{budget} & Sparsification according to an overall target budget (Eq.~\ref{eq:budget}).\\
\hline
\end{tabular}
\end{table}
\setlength{\tabcolsep}{1.4pt}

\begin{table}[t]
\centering
\caption{Budget-aware pruning approaches under 85\% target sparsity constraint for  WRN-16-8. Both acquired overall sparsity and accuracy are reported}
\label{table:cifar100}
\begin{tabular}{lccc}
\hline\noalign{\smallskip}
& \#Params & Sparsity & Acc.\\
Method & (Million) & (\%) & (\%) \\
\noalign{\smallskip}
\hline
\noalign{\smallskip}
WRN-16-8  & 11.012 & 0 & 78.52 \\ (reference model) & & &\\
WRN-16-3 & 1.672 & 85.00 & 74.89 \\ (equivalent dense model) & & &\\
\hline 
Proposed-Fixed & & & \\
\hline
fixed-BS & 1.719 & 84.90 & 78.66 \\
fixed-GA & 1.588 & 86.09 & 78.45 \\
\hline
Proposed-Adaptive & & & \\
\hline
budget-avg & 2.559 & 77.21 & 78.62 \\
budget-wavg & 1.655 & 85.45 & 78.55 \\
\hline
\end{tabular}
\end{table}

\begin{figure}
\centering
\includegraphics[width=0.65\linewidth]{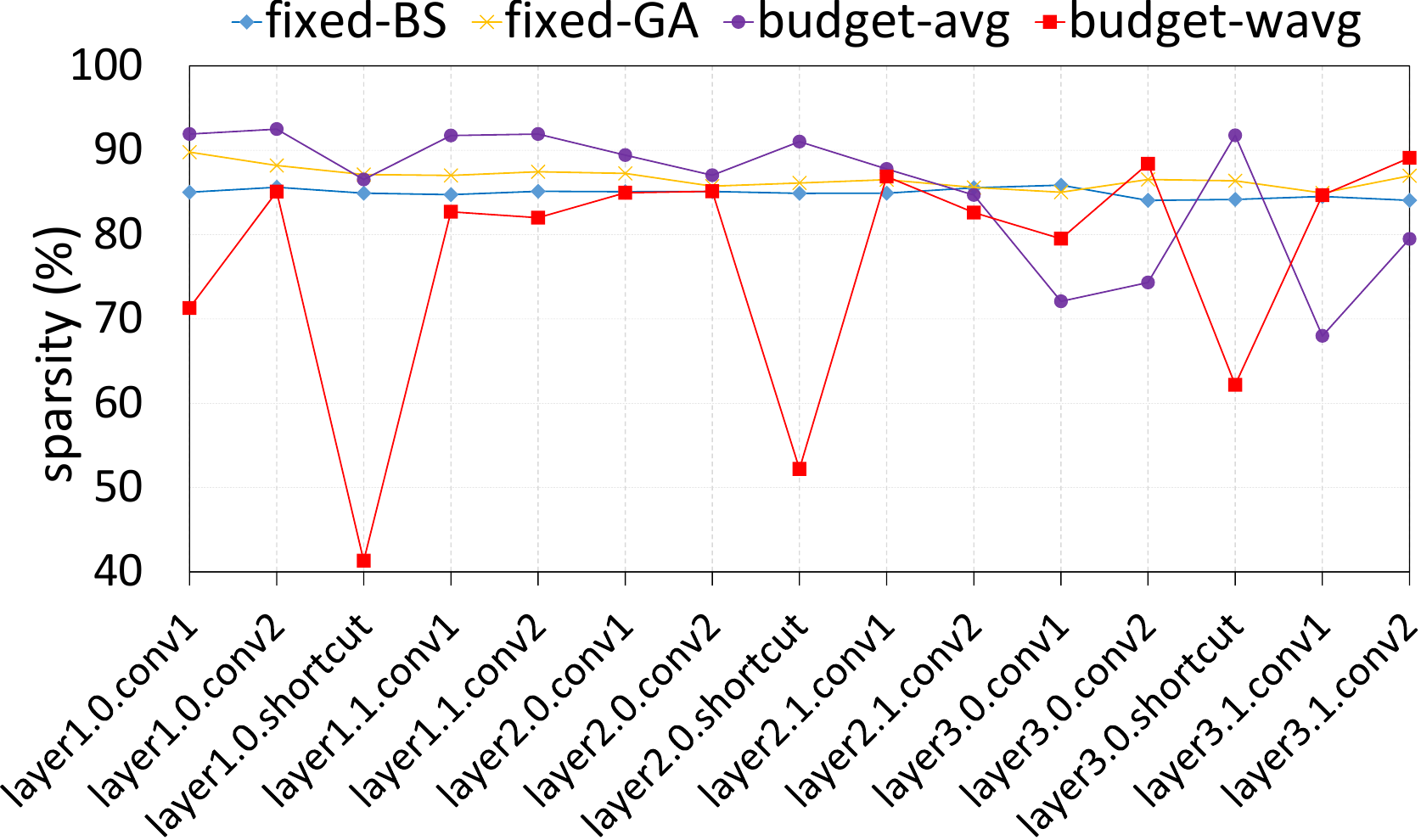}
\caption{Actual per layer sparsity for WRN-16-8 under the constraint of 85\% overall sparsity.}
\label{fig:per_layer_sparsity}
\end{figure}

Notably, even if a considerable amount of parameters are pruned (nearly 10 million), no accuracy loss is observed, contrary to the equivalent dense model of the same size, while all the reported pruning approaches have similar performance to the initial over-redundant model.
Moreover, the binary search approach presents the best approximation of the target sparsity, confirming its superior precision compared to the Gaussian assumption alternative. Nevertheless, both Gaussian assumption alternatives display a minor sparsity divergence which is not important when considering practical budget-constrained applications. 
More importantly, we can conclude that the average sparsity constraint (\emph{budget-avg}) is not appropriate for budget-constrained tasks, since it does not take into account the difference in number of parameters between different layers and results to an overall sparsity of $77.21\%$.
On the contrary the weighted average alternative is built to address this problem and thus generates a precise overall sparsity (\emph{wavg} would be considered as the default adaptive pruning approach for the rest of the paper).
The effect of these methods on the per layer sparsity distribution is presented at Figure~\ref{fig:per_layer_sparsity}. As expected, the binary search approach achieves the target sparsity for all layers.
The Gaussian assumption variant has the same behavior, similar to binary search yet less precise, while the network-wise adaptive variants display significant fluctuations.
Specifically, the average sparsity alternative displays comparatively low sparsity at the last conv layers, which leads to the decrease of the overall sparsity, whilst weighted average alternative displays high compression on the last layers and notably low at shortcut layers. This result is aligned with our initial goal of relaxed, adaptive sparsification per layer.

\begin{figure}
\centering
\includegraphics[width=0.55\linewidth]{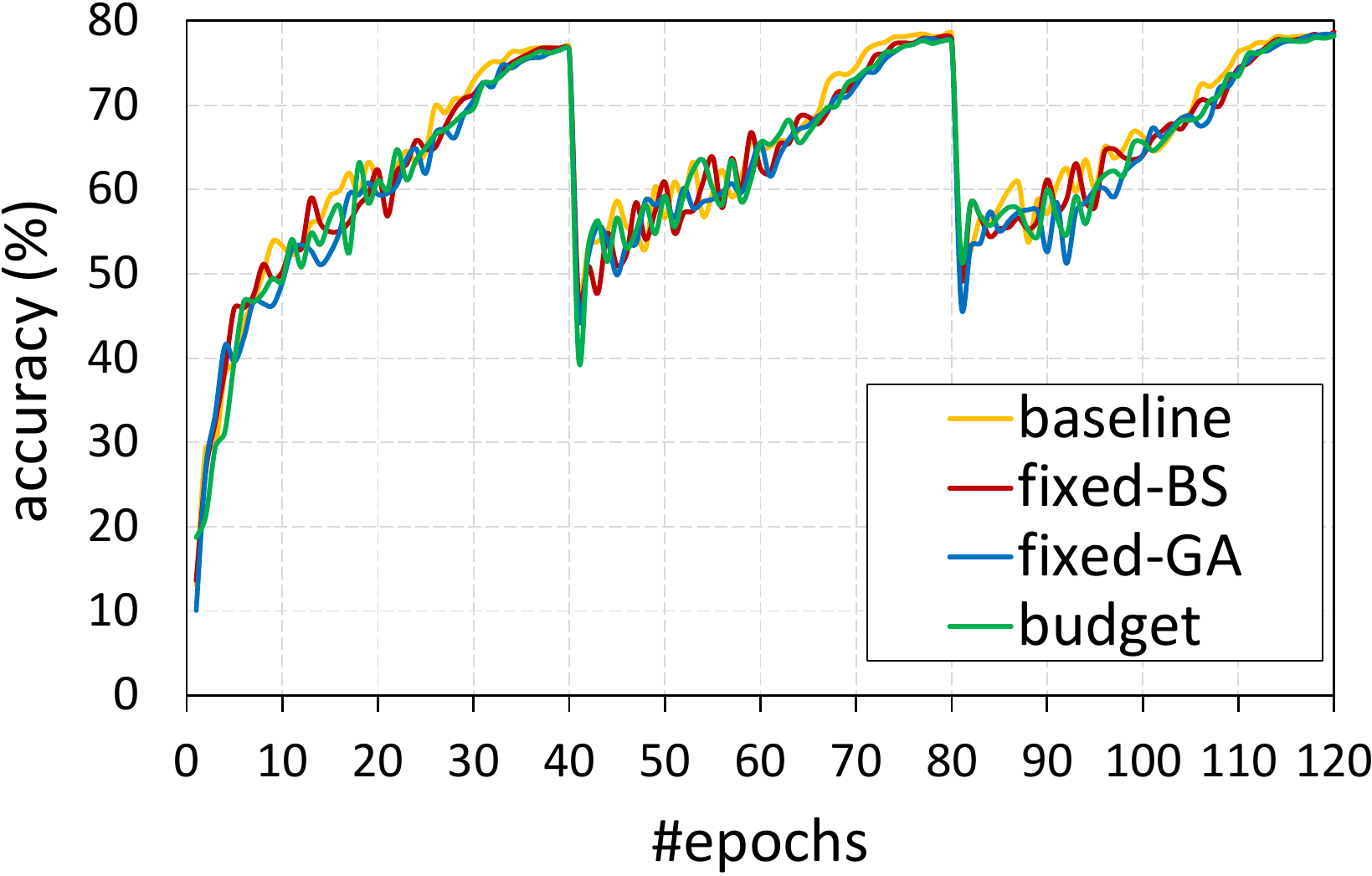}
\caption{Convergence of the different pruning alternatives under the constraint of 85\% overall sparsity (WRN-16-8 / CIFAR-100).}
\label{fig:convergence}
\end{figure}

Regarding the convergence rate of our methods, we report the evolution of the evaluated model's accuracy over training in Figure~\ref{fig:convergence}. We can observe that, indeed, the convergence of the proposed methods is akin to that of the initial dense uncompressed network, as we have already advocated in the analysis of Section~\ref{sec:proposed}.

\begin{figure}
\centering
\includegraphics[width=0.55\linewidth]{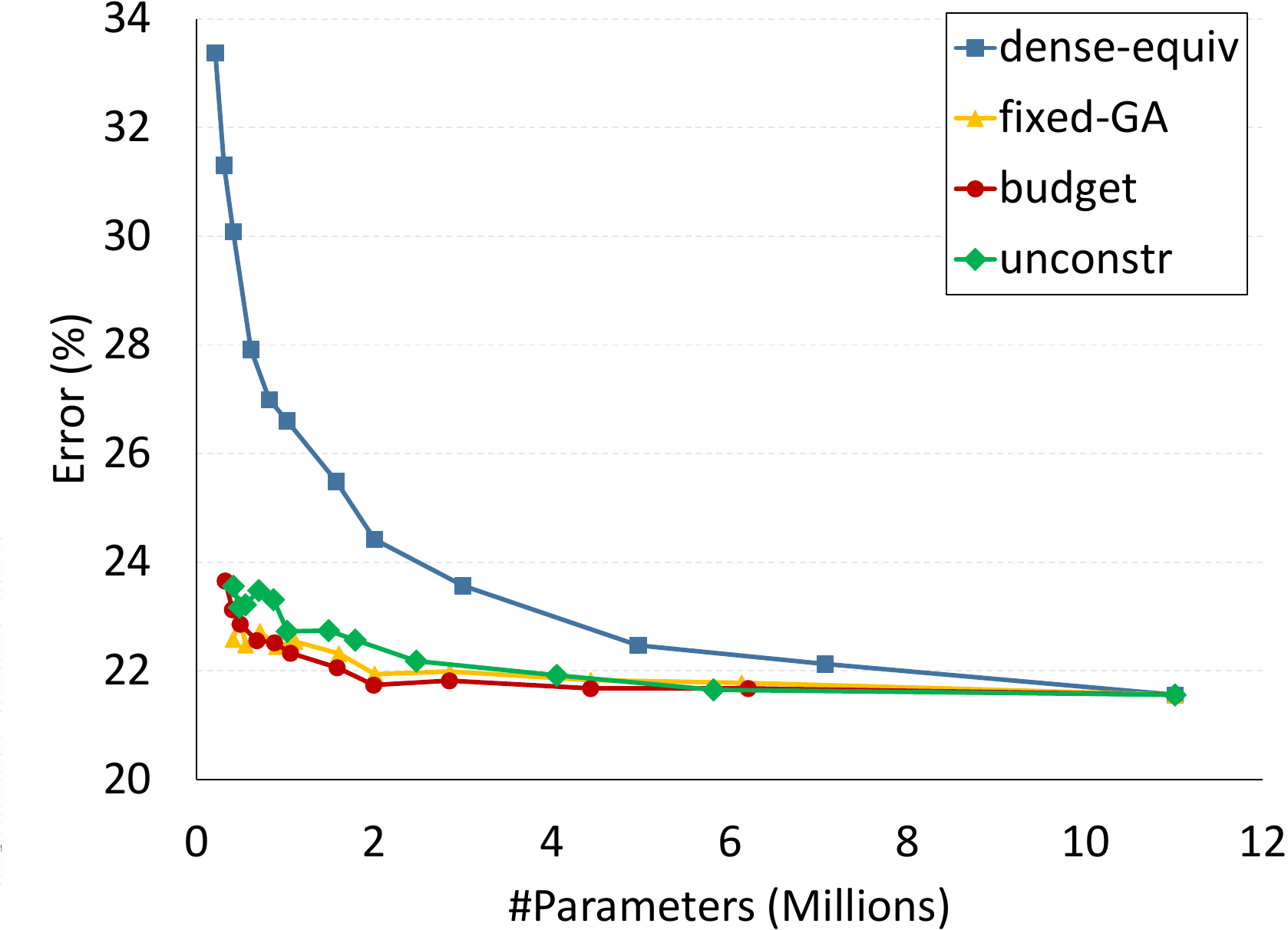}
\caption{Trade-off curves for the proposed pruning approaches and the dense equivalent models (WRN-16-8 / CIFAR-100).}
\label{fig:cifar-curves}
\end{figure}

To thoroughly evaluate the proposed methods, we use the analysis of Section~\ref{sec:fair_comparison} to generate the error vs number of parameters plot for both the \emph{fixed-GA} and \emph{budget} approaches, while we also consider the unconstrained alternative (Eq.~\ref{eq:wavg}), referred to as \emph{unconstr} (we select a set of $\lambda$ sampled in $[0.01, 10.0]$, since different values correspond to different acquired sparsity levels). 
The generated curves are shown at Figure~\ref{fig:cifar-curves}, where we can notice the effectiveness of the similarly performing proposed methods compared to the dense equivalent baseline.

\subsection{ImageNet}

Due to resource limitations, experiments on ImageNet were performed by sparsifying existing pre-trained networks\footnote{taken from PyTorch model zoo} for only 20 epochs. 
First, we consider the AlexNet architecture, aiming to minimize the overall size of the network without sacrificing accuracy (a popular setting for pruning techniques).
To this end, we apply the unconstrained variation ($\lambda=10$), which is designed for the task of compressing a model with minimal accuracy loss.
AlexNet consists of 8 layers (5 convolutional and 3 fully-connected) with high variability in size, which is ideal to showcase the value of adaptive pruning approaches. 
Table~\ref{table:imagenet} compares the results of our method with existing unstructured pruning techniques and shows we achieve significant compression with no accuracy loss (on par with the state-of-the-art frequency domain pruning approach BA-FDNP~\cite{NIPS2018_7382}).          

\setlength{\tabcolsep}{4pt}
\begin{table}
\centering
\caption{Comparison of several unstructured pruning approaches on AlexNet/ImageNet setting.}
\begin{tabular}{lrcc}
\hline\noalign{\smallskip}
& \#Params & Top-1 Acc. & Top-5 Acc.\\
Method & (Million) & (\%) & (\%) \\
\noalign{\smallskip}
\hline
\noalign{\smallskip}
AlexNet reference & 61.10 & 56.58 & 79.88 \\
\hline
LWC~\cite{NIPS2015_5784} & 6.71 & 57.20 & 80.30\\ 
L-OBS~\cite{NIPS2017_7071} & 6.71 & 56.89 & 79.99  \\ 
DNS~\cite{NIPS2016_6165} & 3.48 & 56.91 & 80.01 \\ 
Constraint-aware~\cite{10.1007/978-3-030-01237-3_25} & 2.97 & 54.84 & - \\ 
ADMM~\cite{10.1007/978-3-030-01237-3_12} & 2.90 & - & 80.20 \\ 
FDNP~\cite{NIPS2018_7382} & 2.90 & 56.84  & 80.02 \\ 
BA-FDNP~\cite{NIPS2018_7382} & 2.70 & 56.82 & 79.96 \\ 
\hline
unconstrained & 2.65 & 56.62 & 79.64\\
\hline
\end{tabular}
\label{table:imagenet}
\end{table}
\setlength{\tabcolsep}{1.4pt}

\begin{figure}
\centering
\includegraphics[width=0.65\linewidth]{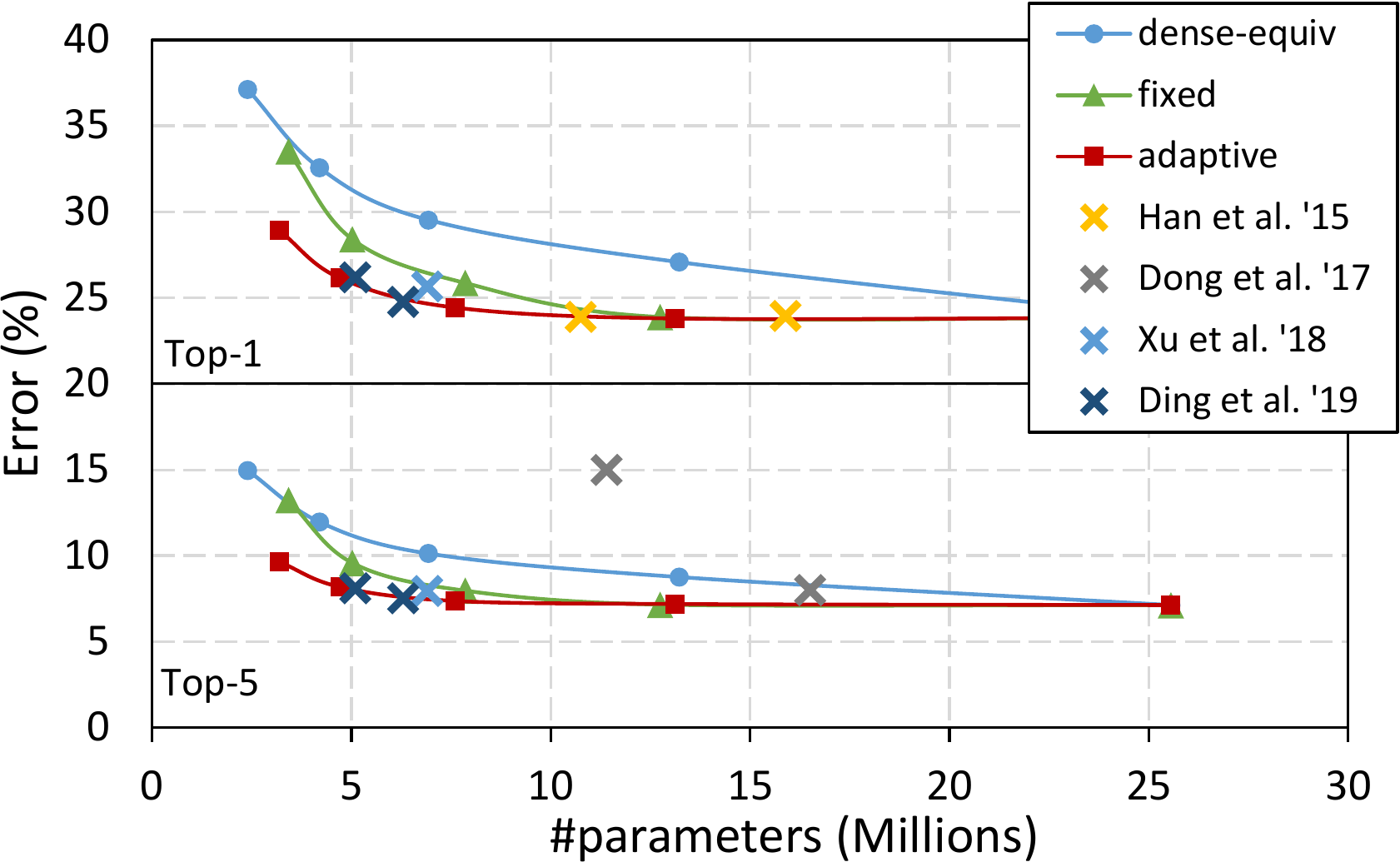}
\caption{Trade-off curves of the proposed variations on ResNet50/ImageNet setting along with several SoA unstructured pruning approaches.}
\label{fig:imagenet}
\end{figure}

Next, we consider the case of ResNet50 for the proposed fixed (GA) and adaptive budget-constraint variants, as shown at Figure~\ref{fig:imagenet}. 
Budget-aware variants are more suitable to construct such trade-off curves, since we can specifically define the requested compression, and thus we do not evaluate the unrestrained variant on this setting.
The dense equivalent models for the trade-off curve were trained from scratch for 60 epochs, which is not optimal, but provides a credible accuracy vs compression trend.
In this setting the flexibility and superiority of the adaptive (budget-constrained) approach is evident regardless the requested sparsity.
On the contrary the fixed sparsity approach does not perform well when high compression rates are considered and it has similar performance to the dense equivalent baseline.
Compared to SoA approaches, our adaptive pruning alternative has better accuracy for the same number of parameters, except the very recent work of Ding et al.~\cite{ding2019global}, which has similar performance.
Note that, based on the formed trade-off trends of Figure~\ref{table:imagenet}b, we can increase the overall sparsity up to 70\% ($\sim$7.5M parameters, i.e. smaller than the dense ResNet18 architecture) without any significant accuracy loss.

\section{Conclusions}

In this paper, we have presented minimal-overhead budget-aware compressive learning algorithms for neural networks.
We first established two efficient sparsity controlling algorithms for pruning individual layers given a specific predefined sparsity constraint. 
Next, we extended these algorithms into a novel intuitive sparsity loss, ideal for adaptive budget-aware optimization over the whole network. 
The proposed adaptive sparsity loss efficiently formulates each layer's sparsity as a differentiable function by adding a single trainable parameter per layer.  
Extensive experiments demonstrate the effectiveness of the proposed pruning methods for the image classification task on different datasets and network architectures.

\clearpage
\bibliographystyle{unsrt}  
\bibliography{sparse_arxiv}

\clearpage
\appendix
\section*{Appendix: Back-propagating through STE with Adaptive Sparsity Loss}

Adaptive sparsity loss relies on a trainable parameter $b$ (per-layer), which denotes the bound of a magnitude-based pruning operation. 
The task-related loss is affected by these bounds even if we do not add the proposed sparsity loss ($\lambda=0$), i.e. there exists a gradient flow from the task-related loss to the bounds.
Intuitively, if no sparsity controlling loss exists, the bounds, and consequently the per-layer sparsity, should be minimized in order to use as much weights as possible, unhindered.
In this section, we will show the effect of STE to a bound parameter $b$ by deriving its corresponding gradient.

First, we re-define the pruning threshold function $f_{prune}$ with a fixed unity boundary as:
\begin{equation}
    T(x) =
    \begin{cases}
      0, & \text{if}\ |x| < 1 \\
      x, & \text{otherwise}
    \end{cases}
\end{equation}
\sloppy We denote the vectorized version of the weights and the pruned weights as $\text{vec}(\mathbf{W}) = [w_0, w_1, \dots w_{n-1}]$ and $\text{vec}(\widetilde{\mathbf{W}}) = [\widetilde{w}_0, \widetilde{w}_1, \dots \widetilde{w}_{n-1}]$, respectively.
Then, using the Gaussian Assumption approach (given a zero mean value), a weight $w_i$ is pruned according to:
\begin{equation}
    \widetilde{w}_i =
    \begin{cases}
      0, & \text{if}\ |w_i| < b \sigma \\
      w_i, & \text{otherwise}
    \end{cases}
\end{equation}
Or, equivalently, using $T(\cdot)$:
\begin{equation}
\widetilde{w}_i = b \sigma T\Big(\frac{w_i}{b \sigma}\Big)
\end{equation}

Both $\mathbf{W}$ and $b$ are trainable parameters and therefore we need to define their respective gradients. 
To compute the task-related loss ($L$) gradient w.r.t. the pruned weights ($\vartheta L / \vartheta \widetilde{\mathbf{W}}$), we follow the typical formulation of neural networks and their back-propagation process. Nevertheless, in order to fully describe the training process of the proposed work, we should define the derivatives $\vartheta \widetilde{w}_i / \vartheta w_i$ and $\vartheta \widetilde{w}_i / \vartheta b$.
Note that STE is applied on the thresholding function $T(x)$, i.e. we assume the identity function during the backpropagation.

The first case of $\vartheta \widetilde{w}_i / \vartheta w_i$ is rather straightforward:
\begin{align}
   \frac{\vartheta \widetilde{w}_i}{\vartheta w_i} = \frac{\vartheta \bigg( b \sigma T\Big(\frac{w_i}{b \sigma}\Big)\bigg)} {\vartheta w_i} \stackrel{\text{\emph{STE}}}{=} \frac{b \sigma \vartheta   \Big(\frac{w_i}{b \sigma}\Big)} {\vartheta w_i} = 1
\end{align}
This is inline with the main functionality of STE, which passes through the gradient to the weights as if no threshold operation was applied.
However, the gradient of a pruned weight $\widetilde{w}_i$ with respect to the single extra parameter $b$, that control sparsity, has not such simple and intuitive interpretation.
Specifically, the gradient is computed as:
\begin{align}
   \frac{\vartheta \widetilde{w}_i}{\vartheta b} & = \frac{\vartheta \bigg( b \sigma T\Big(\cfrac{w_i}{b \sigma}\Big)\bigg)} {\vartheta b} 
   = \sigma T\Big(\frac{w_i}{b \sigma}\Big) + b \sigma \frac{\vartheta T\Big(\cfrac{w_i}{b \sigma}\Big)} {\vartheta b} \nonumber\\
   & \stackrel{\text{\emph{STE}}}{=} \sigma T\Big(\frac{w_i}{b \sigma}\Big) + b \sigma\frac{\vartheta \Big(\cfrac{w_i}{b \sigma}\Big)} {\vartheta b}
   = \frac{1}{b} \widetilde{w}_i - \frac{1}{b} w_i
\end{align}
Overall, the loss gradient with respect to $b$ is computed as:
\begin{align}
   \frac{\vartheta L}{\vartheta b} = \sum_{i=0}^{n-1}\Big( \frac{\widetilde{w}_i-w_i}{b} \Big) \frac{\vartheta L}{\vartheta \widetilde{w}_i}
\end{align}

The derived gradient has some notable properties:
\begin{itemize}
    \item The gradient is minimized when either the majority of the weights is unpruned (which corresponds to a low bound $b$) or a well-performing model is found and no further action is needed ($\vartheta L / \vartheta \widetilde{w}_i \to 0$). Note that only pruned weights ($\widetilde{w}_i=0$) contribute to the gradient.
    \item If STE was not applied, the gradient would be zero. This behavior is not desirable for the proposed sparsity controlling problem. In fact, the bound minimization property under the absence of a sparsity inducing loss is very intuitive and in line with the abstract idea of a robust sparsification approach.
    
\end{itemize}

Regarding the sparsity loss, its gradient (for a single layer) $L_s = 1 - \text{erf}(b / (\sigma \sqrt{2}))$ with respect to $b$ is: 
\begin{align}
   \frac{\vartheta L_s}{\vartheta b} = -\frac{2 e^{-b^2 / (2\sigma^2)}}{ \sigma \sqrt{2 \pi}}
\end{align}

\noindent \textbf{Remark:} Even though we defined the pruning function (and consequently the losses) with respect to both boundary $b$ and standard deviation $\sigma$, we did not take into account that $\sigma$ is also a function of the layer's weights, i.e.
$\sigma = f(\mathbf{W})$, and thus, theoretically, we can also back-propagate the derived error towards the weights. 
However, we choose not to propagate through $\sigma$ to avoid adding extra gradient flows that would over-complicate training.
In fact, we consider $\sigma$ as an intrinsic, non-adjustable, property of the weight distribution and thus we control the pruning operation only through the trainable threshold $b$.

\end{document}